\documentclass{article}
\usepackage{ijcai26}

\usepackage{times}
\usepackage{soul}
\usepackage{url}
\usepackage[hidelinks]{hyperref}
\usepackage[utf8]{inputenc}
\usepackage[small]{caption}
\usepackage{graphicx}
\usepackage{amsmath}
\usepackage{amsthm}
\usepackage{booktabs}
\usepackage{algorithm}
\usepackage{algorithmic}
\usepackage[switch]{lineno}
\usepackage{xcolor}

\title{A Unified Moral-Value Dataset for Instruction Tuning}

\author{
Zhaohui Zeng$^1$
\and
Florian Mai$^{2,3}$\\
\affiliations
$^1$RWTH Aachen\\
$^2$University of Bonn\\
$^3$Lamarr Institute for Machine Learning and Artificial Intelligence\\
\emails
zhaohui.zeng@rwth-aachen.de,
fmai@uni-bonn.de
}
\begin{document}

\maketitle

\begin{abstract}
    Large language models (LLMs) have developed rapidly and become valuable tools in everyday life. However, how to align LLMs to a particular set of human values is still an open problem. Recent studies show that instruction tuning has strong potential for zero-shot tasks and may serve as an effective approach to addressing value alignment. Nevertheless, although many datasets for instruction tuning already exist, they are not specifically designed around moral scenarios and behaviors. We construct a unified moral-value dataset that can be directly used for instruction tuning. This dataset is built upon existing moral-value datasets by merging them into a unified corpus and converting them into an instruction–response format. We show that training on a mixed dataset combining general task datasets with our dataset preserves general-task performance, and we report preliminary observations on how the mixing ratio affects value-oriented task performance. Our work provides a moral-value dataset for instruction tuning and offers a useful resource for further alignment research. The dataset is available at \url{https://huggingface.co/datasets/teohzzh/value-for-instruction-tuning}.
\end{abstract}

\section{Introduction}

In recent years, large language models (LLMs) have become a central research focus in natural language processing and artificial general intelligence (AGI). From early pre-trained language models~\cite{radford2018improving,devlin-etal-2019-bert} to more recent advanced models~\cite{brwon2020gpt3,touvron2023llamaopenefficientfoundation}, LLMs have demonstrated strong versatility in tasks such as text generation, question answering, translation, and decision support for complex problems~\cite{wei2022chainofthought,bubeck2023sparksartificialgeneralintelligence}. 

However, despite the success of current LLMs, there are still many limitations and challenges. LLMs lack stable controllability when generating content, making it difficult to consistently adhere to human intent or safety guidelines. Consequently, this alignment issue can lead LLMs to generate unexpected outputs, and, in some cases, malicious exploitation to disseminate false information or harmful content~\cite{bender2021onthedangers}. The central challenges in value alignment research lie in understanding the intricacies inherent in human value systems and mapping these values onto the behavior of AI systems. For the former, human values are generally understood as the fundamental principles that guide individual judgment and behavioral choices. For many years, researchers have been trying to summarize and categorize human values, leading to two main approaches: philosophy such as normative ethics~\cite{Deigh_2025} and psychology such as Moral Foundations Theory~\cite{brandt2013morality}.
For the latter, existing research typically describes the alignment process as aligning the model output with human preferences, social norms, or ethical principles through specific training strategies~\cite{amodei2016concreteproblemsaisafety}. Nevertheless, due to the diversity and context-dependent nature of human values, effective alignment is difficult to achieve through a single method and often relies on a combination of multiple technological approaches.

Besides training strategies directly designed for alignment, recent research has shown that instruction tuning also plays an important role in improving the model's generalization and instruction-following capabilities~\cite{WeiBZGYLDDL22}. Therefore, although instruction tuning was not initially proposed to solve the alignment problem, it still provides a potential approach to improve the model's value alignment. However, although numerous instruction tuning datasets exist~\cite{openasistent,lambert2024tulu3}, none are specifically designed to address the value alignment problem. As a result, a gap remains. On the one hand, instruction tuning provides a flexible training paradigm. On the other hand, effectively leveraging diverse value-oriented resources remains challenging. Bridging this gap requires a systematic approach that can unify heterogeneous value datasets and incorporate them effectively into instruction tuning.

To tackle this challenge, we develop a unified pipeline that integrates multiple moral-value datasets into instruction tuning. By transforming heterogeneous datasets into a consistent instruction format and incorporating them into the training process, the model can learn value-aware behaviors. At the same time, it maintains strong general task performance. 

In summary, our contributions are listed as follows:

\begin{itemize}
    \item We review and collect three moral-value datasets and integrate them into a unified dataset with a consistent format and schema. We then train a ModernBERT model to impute missing values. 
    \item We design prompting strategies to generate instruction–response templates and combine them with the unified dataset to construct a moral-value instruction tuning dataset.
    \item We fine-tune large language models on our moral-value instruction tuning dataset under different data mixing ratios with a regular dataset. We then evaluate the resulting models, observing that general-task performance is preserved across configurations and that the mixing ratio influences performance on value alignment tasks.
\end{itemize}

\section{Related Works}
\subsection{Value Alignment}
Existing value alignment methods can be broadly categorized into two types: rule-based methods and preference-based methods. 

Rule-based alignment methods introduce explicit behavioral norms or ethical guidelines to constrain model outputs. A typical rule-based method is constitutional AI. It proposes a representative framework that guides the model to adhere to specific value norms during generation and self-evaluation by specifying a set of constitutional principles, without relying on large amounts of manually labeled data~\cite{bai2022constitutionalaiharmlessnessai}. 

Unlike rule-based alignment methods, another mainstream direction is to align models using human-generated data. A common approach is supervised fine-tuning (SFT), where models are trained on curated data that explicitly demonstrate desired behaviors~\cite{ouyang2022training}. Instead of learning preferences implicitly, SFT directly teaches the model how to respond in specific situations. Building on this, preference-based alignment methods further refine model behavior by learning from human preferences over model outputs. The most widely used approach is reinforcement learning with human feedback (RLHF), which constructs a reward model based on human preferences and combines it with reinforcement learning optimization~\cite{christiano2017deep,ziegler2020finetuninglanguagemodelshuman,ouyang2022training}. In practice, SFT is often used as a preliminary step in RLHF, providing an initial aligned model before preference-based optimization ~\cite{ouyang2022training}. Subsequently, researchers proposed several improvement methods. For example, Direct Preference Optimization (DPO) improves training stability and efficiency by directly optimizing the distribution of preference data, thus avoiding complex reinforcement learning processes~\cite{rafailov2023direct}. Preference-based methods demonstrate strong flexibility and adaptability in practice, but their effectiveness is highly dependent on the quality and distribution of the data, and they may inherit or amplify biases present in human annotations~\cite{sharma2025understandingsycophancylanguagemodels}.

\subsection{Instruction Tuning}

Instruction Tuning is a training method that improves a model's ability to understand and execute natural language instructions through supervised learning on instruction-response data pairs. Unlike traditional pre-training, which primarily focuses on language modeling objectives based on unlabeled text, instruction tuning introduces explicit task descriptions, enabling the model to handle various task formats within a unified framework~\cite{WeiBZGYLDDL22}. Building on this foundation, instruction tuning has gradually evolved into a problem-solving paradigm that uses natural language instructions as a unified interface, enabling different tasks to be modeled and processed within a unified text-to-text framework~\cite{chatterjee2025onthe}.

\subsection{Moral-Value Datasets}
Value-oriented datasets refer to datasets that include human-annotated moral judgments. They are often derived from specific value frameworks and consist of collected real-world or hypothetical scenarios. The ETHICS dataset is a benchmark designed to evaluate whether machine learning models understand fundamental human moral judgments~\cite{hendrycks2021ethics}. The dataset covers multiple ethical dimensions derived from normative ethics theories, containing more than 130,000 examples. To not only determine whether an action is moral, but also to capture the entire process of moral decision making, the UNIMORAL dataset is designed to model the full pipeline of human moral reasoning~\cite{kumar-jurgens-2025-rules}. SOCIAL-CHEM-101 is a dataset used to study everyday social norms and moral judgments~\cite{forbes2020social}. The dataset's main goal is to understand how people use implicit social norms to make decisions in real-world social situations. To do this, it introduces the concept of Rules-of-Thumb (RoTs). These RoTs represent the social rules people follow in specific situations when evaluating whether a behavior is appropriate, inappropriate, or acceptable.

\section{Methodology}
\subsection{Framework Overview}
\begin{figure*}[t]
\centering
\includegraphics[width=\textwidth]{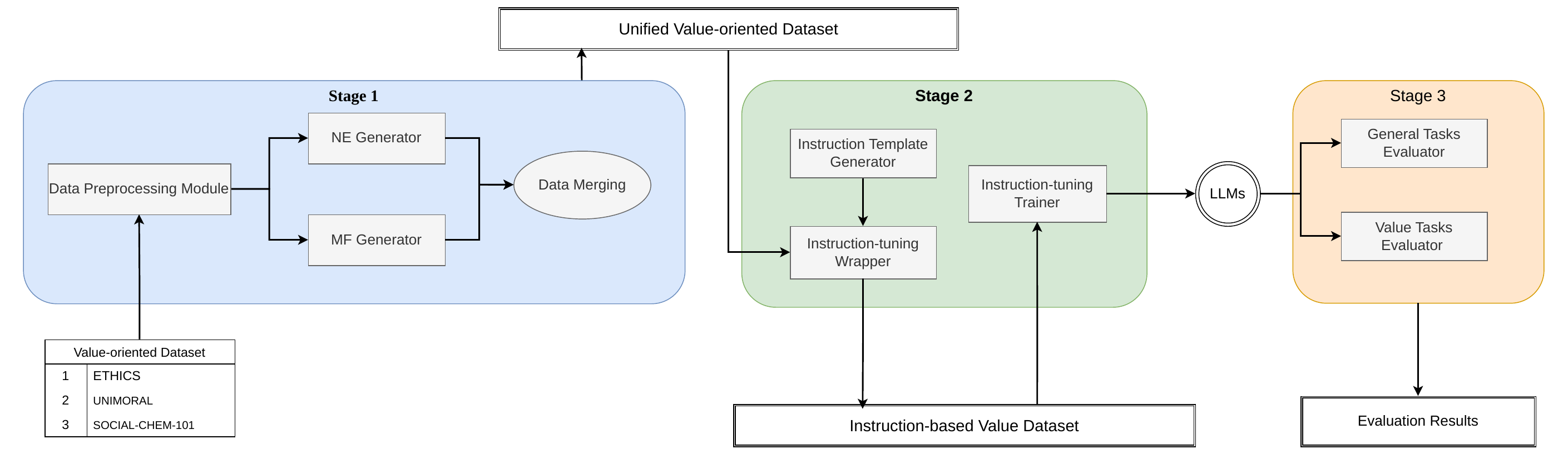}
\caption{Overall pipeline of our proposed dataset construction, instruction tuning, and downstream evaluation.}
\label{fig:frameworkov}
\end{figure*}

To build the unified dataset, we propose a three-stage pipeline. The pipeline is shown in Figure~\ref{fig:frameworkov}.

In Stage 1, we first employ a data preprocessing module to standardize the filtered moral-value datasets into a unified format: \texttt{\{scenario, value\_framework, label\}}. Then, these data are fed into two different data generators for data expansion under two different value frameworks. For datasets based on the normative ethics framework, each scenario is paired with all moral foundation categories and then input into the trained Moral Foundation Theory data generator to obtain the corresponding moral foundation labels. Similarly, for datasets based on the moral foundation theory framework, each scenario is paired with all normative ethics categories and fed into the trained Normative Ethics data generator to produce the corresponding normative ethics labels.
Through this process, all datasets are augmented with two sets of value frameworks and their corresponding labels, which can then be directly integrated in the data merging module.

In Stage 2, prompt engineering is used to guide the LLM in generating a diverse set of templates, including roles, instructions, and responses. With the help of an Instruction-Tuning Wrapper, these templates are randomly combined with the data to construct instruction-tuning datasets. These datasets can then be directly used for fine-tuning using the same open-instruction training pipeline as the Tulu3 model~\cite{lambert2024tulu3}. During training, multiple models can be obtained by adjusting the configuration of the training data. Specifically, we consider different data mixing configurations. The moral-value instruction-tuning dataset is mixed with the regular instruction-tuning dataset using different strategies, resulting in different fine-tuned models.

In Stage 3, we evaluate all fine-tuned models. For general tasks, we adopt the Open Language Model Evaluation System (OLMES)~\cite{gu-etal-2025-olmes} to standardize the benchmarking process, including prompt formatting, few-shot configurations, and scoring procedures. For value-related tasks, we follow the evaluation pipeline of Value-Action Gap~\cite{shen-etal-2025-mind}, which allows us to examine the consistency between a model’s value endorsement and its behavioral choices. This evaluation has two components:

\begin{itemize}
    \item \emph{Task 1: State Value Inclination:} The model is asked to determine how strongly it agrees or disagrees with a given value.
    \item \emph{Task 2: Select Value-Informed Action: } Given a value and a scenario, the model must choose between two possible actions: one that aligns with the specified value and one that does not.
\end{itemize}

After completing both tasks, we examine whether the model exhibits inconsistencies between its stated values and value-informed actions. Such inconsistencies indicate a potential value-action gap. For instance, a model may disagree with a value in Task 1 but select an action aligned with that value in Task 2, or vice versa. By combining the two evaluations, we can better understand how fine-tuning influences the model's behavior.

\subsection{Value Dataset Construction}
In the first stage of our pipeline, the goal is to construct our value dataset from multiple existing moral-value datasets. This process involves several key steps. 

\begin{itemize}
    \item Collect and examine high-quality moral-value datasets to gain a comprehensive understanding of their characteristics.
    \item Build a data preprocessing module. As these datasets were developed with different research objectives and methodologies, they often vary significantly in both structure and data representation. Therefore, we design a data preprocessing module that converts all datasets into a unified format: \texttt{\{scenario, value\_framework, label\}}.
    \item Train and use data generators. Before merging the datasets into a single corpus, there is one additional challenge that needs to be resolved: the difference between the value frameworks used in the original datasets. To tackle this issue, we train data generators to assign additional value annotations for each instance when such annotations are not provided in the original datasets.
\end{itemize}

We select high-quality moral-value datasets according to three criteria: (1) they are publicly available, (2) they provide annotations consisting of scenarios and corresponding moral judgments, and (3) they contain a sufficient number of instances for instruction tuning. After carefully reviewing the available data, we select ETHICS~\cite{hendrycks2021ethics}\footnote{\url{https://huggingface.co/datasets/hendrycks/ethics}}, UNIMORAL~\cite{kumar-jurgens-2025-rules}\footnote{\url{https://huggingface.co/datasets/shivaniku/UniMoral}} and SOCIAL-CHEM-101~\cite{forbes2020social}\footnote{\url{https://huggingface.co/datasets/tasksource/social-chemistry-101}} to construct our unified value dataset.

We first build the data preprocessing module to filter and recombine the three selected datasets based on the characteristics of each dataset, converting them into a unified structure for subsequent merging.

For the ETHICS dataset, most subdatasets already follow the format \texttt{\{scenario, value\_framework, label\}}. However, two categories still need to be processed into this format. In the Utilitarianism subdataset, each original data entry contains two actions: one baseline action and one less pleasant alternative. We first randomly assign these actions to \texttt{Option A} and \texttt{Option B}. Each entry is then expanded into two instances, each selecting one of the options. The instance in which the baseline action is selected is labeled as 1.  For the Commonsense subdataset, a label of 1 denotes morally wrong behavior, while 0 denotes behavior that is not morally wrong, which is the opposite of the convention used in the other subdatasets. Therefore, we flip these labels before merging them with the others. 

For the UNIMORAL dataset, we first perform base text construction, in which the original information spanning multiple fields is concatenated into a single scenario. Each data entry is expanded into two entries, each corresponding to one of the possible actions provided by the original dataset. After concatenation, the data is further expanded to include annotations for different normative ethics categories. In the UNIMORAL dataset, each selected action is annotated according to four normative ethical types. Each type is assigned a rating from 1 to 5 within the \texttt{action\_criteria} field. To obtain binary labels indicating whether an action is morally correct (1) or not (0), we apply a label filtering process to convert these ratings. Specifically, ratings of 1 and 2 are mapped to 0, ratings of 4 and 5 are mapped to 1, and ratings of 3 are discarded.

In the SOCIAL-CHEM-101 dataset, the key fields for our value dataset construction are \texttt{Action}, \texttt{Rot\_judgement}, \texttt{Rot\_moral\_foundations} and \texttt{Action\_agree}.  For each data entry, the \texttt{Action} and \texttt{Rot\_judgement} fields are concatenated to form the scenario during base text construction. Meanwhile, a moral foundation splitting step is performed to separate all \texttt{Rot\_moral\_foundations} associated with the same action. Each scenario is then expanded into multiple entries, one for each moral foundation, while the same \texttt{Rot\_judgement} is assigned as the rating for each resulting entry.

After data preprocessing, the resulting dataset contains the fields \texttt{\{scenario, ethics\_type, ethics\_label, mft\_type, mft\_rating, language, source, n\_chars, split\}}. Due to the nature of the original datasets, some field values are missing and are therefore filled with \texttt{None} during the procedure. For example, data from the ETHICS dataset lack annotations for moral foundation types and corresponding labels, and are thus assigned \texttt{None} in the \texttt{mft\_type} and \texttt{mft\_rating} fields. To address this issue, we introduce two data generators to impute the missing values for normative ethics and moral foundations, respectively.

\begin{figure}[!h]
\centering
\includegraphics[width=0.5\textwidth]{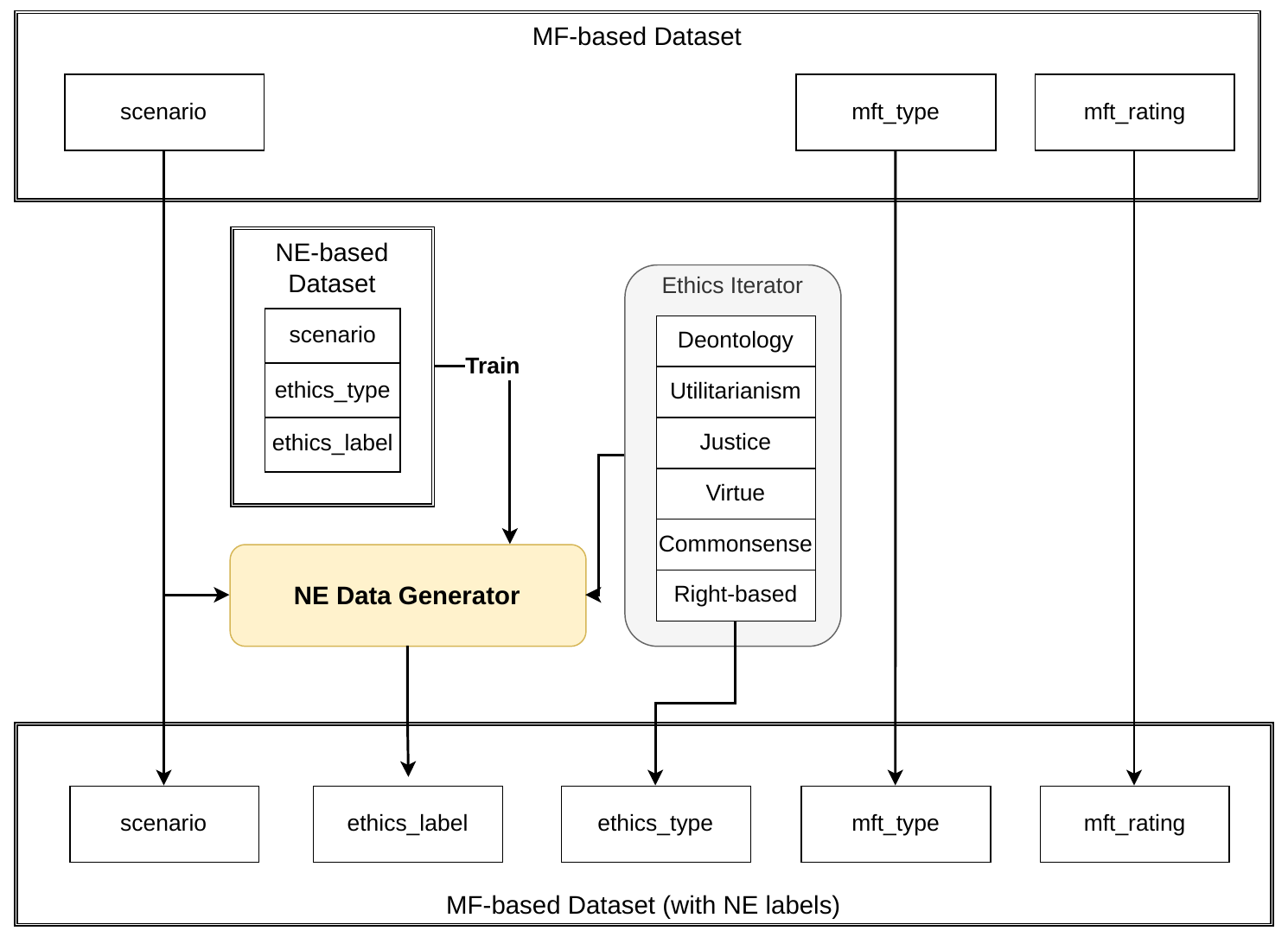}
\caption{Overview of NE data generator. The MF data generator also shares similar structure.}
\label{fig:ne_gen}
\end{figure}

The Normative Ethics (NE) data generator is designed to fill in the missing values in \texttt{ethics\_type} and \texttt{ethics\_label}. This is particularly essential for data originating from datasets annotated solely with moral foundations (e.g., SOCIAL-CHEM-101), where normative ethics information is absent. An overview of the procedure is illustrated in Figure~\ref{fig:ne_gen}.

The input for the NE data generator is a scenario and one normative ethics type (Deontology, Utilitarianism, Justice, Virtue, Commonsense, Right-based). The output is a binary ethics label (1 or 0), indicating whether the scenario is morally correct (1) or wrong (0). For each data instance, we iterate over all ethics types, forming six input pairs with the scenario, and thus generate six new data instances with all fields filled.

A key objective is to build an NE data generator capable of correctly assigning labels to each input pair. To this end, we formulate the problem as a binary classification task, in which the model predicts whether a given scenario is morally acceptable or not. The input is a pair 
$x = (s, t)$, with $s \in \mathcal{S}$ denoting a scenario and 
$t \in \mathcal{T}$ representing an ethics type. 
The goal is to learn a function 
$f: \mathcal{S} \times \mathcal{T} \rightarrow \{0,1\}$ 
that predicts a label $y$. For our NE data generator, model $f$ is a ModernBERT model~\cite{modernbert} trained to solve the binary classification task and the training corpus consists of data from the ETHICS and UNIMORAL datasets.

Similar to the NE data generator, the Moral Foundations (MF) data generator is designed to infer and fill missing values in \texttt{mft\_type} and \texttt{mft\_rating}. Unlike the NE data generator, the MF data generator is not formulated as a binary classification task. Instead, its input consists of a scenario paired with a moral foundation type (e.g., care–harm, fairness–cheating, loyalty–betrayal, authority–subversion, or sanctity–degradation), and the output is a rating ranging from 0 to 4 that reflects the strength of the moral signal. Accordingly, we formulate this task as a multi-class classification problem.  We consider an input pair $x=(s,t)$, where $s \in \mathcal{S}$ denotes a scenario and 
$t \in \mathcal{T}$ representing a moral foundation type. 
The goal is to learn a function 
$f: \mathcal{S} \times \mathcal{T} \rightarrow \{0,1,2,3,4\}$ 
that predicts a label $y$. We also train a ModernBERT model as the MF data generator. Using ModernBERT as the backbone, we experiment with three task formulations for the MF data generator: regression, classification, and ordinal classification.~\cite{Shi2023CORN}.

\subsection{Value-Guided Instruction Tuning}
To be used for instruction tuning, the dataset must be converted into the required instruction-tuning format. In our case, the value dataset still lacks role definitions and instruction prompts. Traditionally, these are manually designed based on specific tasks and data. However, this process is extremely labor-intensive and inefficient. Therefore, we adopt a self-prompting approach~\cite{li2023self-prompt}. The original self-prompting focuses on generating QA pairs and explanations for in-context learning, and has shown that LLMs can effectively bootstrap their own supervision by generating pseudo training examples. Similarly, we further extend this approach by employing LLMs to transform non-instructional data into instruction-response formats by generating instruction templates, thereby converting raw data into instruction-tuning datasets. Specifically, we design a prompt that instructs GPT-5.2 to generate prompt templates in JSON format, according to the principles of prompt engineering and instruction tuning~\cite{reynolds2021prompt}. Each principle is encoded as a constraint in the final prompt for generating instruction–response templates. These generated templates are then combined with our dataset to construct an instruction-tuning dataset. 

Once the instruction-tuning dataset is obtained, we adopt the instruction training pipeline used to train the TULU3 model~\cite{lambert2024tulu3}\footnote{\url{https://github.com/allenai/open-instruct}} as the training framework for fine-tuning on our value dataset. We evaluate several training methodologies provided by the pipeline and selected Supervised Fine-Tuning (SFT) as the primary approach, as it enables the model to learn directly from labeled data, which is well suited to our labeled dataset.

To evaluate the training process, we add an additional step to the original training pipeline. Before training starts, we split the dataset so that a small portion serves as the validation set. During training, we then calculate the loss on the validation set at specific steps.

Furthermore, it is often observed that fine-tuning a model on a specific dataset can degrade its performance on more general tasks. Therefore, instead of training solely on the value dataset, we mix it with a dataset covering general tasks, namely the TULU-3 SFT dataset. By varying the mixing ratio and comparing performance across different downstream tasks, we can evaluate the impact of the value dataset.

\subsection{Evaluation}\label{sec:evaluation}
To evaluate the model's performance on general tasks, we adopt the full OLMES benchmark, including ARC-Challenge, ARC-Easy, BoolQ, CommonsenseQA, HellaSwag, MMLU, OpenBookQA, PIQA, Social IQA, and WinoGrande~\cite{gu-etal-2025-olmes}\footnote{\url{https://github.com/allenai/olmes}}. These tasks cover a diverse range of abilities, including scientific reasoning, commonsense understanding, factual knowledge, and language understanding.

For the evaluation of moral tasks, we adopt the Value-Action Gap benchmark~\cite{shen-etal-2025-mind}\footnote{\url{https://github.com/huashen218/value_action_gap}}. The evaluation of the value-action gap includes two tasks: one to assess the model’s moral inclination and the other to evaluate its behavior. After completing these two tasks, we can not only evaluate how the training data affects the model's moral inclinations and behavior, but also examine whether there is a value-action gap between these two aspects and how the training dataset affects the size of this gap.

When applying the value-action gap pipeline, we enhance its structure using vLLM ~\cite{Kwon2023vllm}\footnote{\url{https://github.com/vllm-project/vllm}} to address the following challenges:

\begin{itemize}
    \item \textbf{Structured Output:} In the value-action gap pipeline, both tasks require standardized JSON outputs. However, the original implementation relies solely on input prompts to constrain the model, which can easily lead to unexpected outputs, especially when using weaker models. Therefore, we adopt vLLM with structured output constraints, constraining the model to generate specific tokens at designated positions to ensure the output adheres to the correct format.
    \item \textbf{Batch Inference:} The original implementation handles one input at a time, leading to high inference latency, especially because the tasks require testing several different prompts for each test instance to rule out the influence of different prompts. We adopt vLLM with batch processing, allowing us to process data in batches and significantly improve inference efficiency.
\end{itemize}

\section{Experiments}
\subsection{Data Generators}
We first evaluate whether the NE and MF data generators can reliably infer missing value annotations before they are used to construct the unified dataset. For the NE data generator, we adopt the pre-trained ModernBERT-large as the backbone model and fine-tune it for a binary classification task. The model predicts whether a given scenario is morally acceptable or morally wrong, corresponding to two output classes. To incorporate ethical context, we reformulate the task as a paired-input setting in Natural Language Inference (NLI) style. Specifically, each input is transformed into a pair consisting of the original scenario as the premise and a prompt as the hypothesis. The prompt explicitly instructs the model to evaluate the scenario under a given ethical framework (e.g., deontology, utilitarianism). Table~\ref{tab:category-results-main} presents the results for each ethics category at the same fixed threshold ($t=0.5$). 

\begin{table}[!h]
    \centering
    \begin{tabular}{lcccc}
        \toprule
        Category & Count & F1 & AP & ECE \\
        \midrule
        Commonsense   & 2,087 & 0.7946 & 0.8866 & 0.1143 \\
        Deontology    & 2,725 & 0.9212 & 0.9748 & 0.0516 \\
        Justice       & 3,269 & 0.8773 & 0.9539 & 0.0860 \\
        Utilitarianism & 4,122 & 0.9168 & 0.9734 & 0.0667 \\
        Virtue        & 4,237 & 0.9738 & 0.9169 & 0.0206 \\
        Overall     & 16,440 & 0.9093   & 0.9540 & 0.0608\\
        \bottomrule
    \end{tabular}
    \caption{Results by ethics category (threshold $t=0.5$). Metrics include F1 score, Average Precision (AP), and Expected Calibration Error (ECE).}
    \label{tab:category-results-main}
\end{table}

The results show clear differences across ethical dimensions. The model performs best on the \textit{virtue} category (F1 = 0.9738), followed by \textit{deontology} and \textit{utilitarianism}. In contrast, the \textit{commonsense} category yields the lowest performance (F1 = 0.7946) and the highest calibration error, indicating that it is the most challenging task.

For the MF data generator, we build upon the same backbone as the NE data generator for all experiments. Depending on the formulation, the model is adapted to perform regression, standard classification, or ordinal prediction over moral judgment scores. We follow the same optimization setup as the NE data generator. The experimental settings are listed in Table~\ref{tab:mfgenerator_config}. For classification-based settings, we primarily rely on cross-entropy loss, while focal loss is optionally introduced to emphasize hard or underrepresented examples. In the ordinal setting, we employ the CORN loss to explicitly capture the ordered structure of labels. These approaches can be further combined with weighting schemes to improve robustness under skewed label distributions. In practice, we assign class weights [2.75, 1.29, 0.46, 0.22, 0.28] based on the distribution of the training set.

\begin{table}[!h]
\centering
\small
\begin{tabular}{lccc}
\toprule
Config & Task Type & Loss & Special Setting \\
\midrule
E2.1 & Regression & MSE & -- \\
E2.2 & Classification & CE & -- \\
E2.3 & Ordinal & CORN & --  \\
E2.4 & Classification & Focal & $\gamma=2.0$  \\
E2.5 & Classification & CE & Class weights  \\
E2.6 & Ordinal & CORN + Focal & $\gamma=2.0$  \\
E2.7 & Ordinal & CORN & Class weights \\
\bottomrule
\end{tabular}
\caption{Training configurations for the MF data generator, MSE denotes Mean Square Error, CE denotes Cross-Entropy}
\label{tab:mfgenerator_config}
\end{table}

The results are summarized in Table~\ref{tab:mfgenerator_results}. The ordinal model (E2.3) achieves the best overall performance in terms of weighted F1 score (0.5334), indicating that explicitly modeling the ordinal structure of the labels is beneficial. In contrast, the regression model (E2.1) achieves the second lowest QWK (0.2574) and the highest within-one accuracy (0.9788), suggesting that regression is effective at predicting values close to the ground truth but less effective at exact classification. Compared to the baseline classification model (E2.2), all ordinal configurations (E2.3, E2.6, E2.7) achieve higher F1 scores and QWK, confirming that ordinal modeling better captures the relative ordering of labels. For classification models, focal loss (E2.4) and class weighting (E2.5) lead to improvements in QWK, with E2.4 achieving the highest QWK (0.3733),
but all models clear the 0.2 threshold for fair agreement~\cite{landis1977measurement}.
However, these improvements are not consistently reflected in F1 or accuracy, suggesting that these methods mainly improve ordinal consistency rather than overall classification performance. Across all configurations, within-one accuracy remains consistently high (above 0.94), indicating that most errors occur between adjacent classes. This suggests that the task is inherently ordinal and that models generally learn the relative ranking even when exact predictions are incorrect.
We therefore adopt the ordinal model (E2.3) as the final configuration for the MF data generator.

\begin{table}[!t]
    \centering
    \begin{tabular}{lcccc}
        \toprule
        Config & F1 & Accuracy & Within-1 Acc & QWK \\
        \midrule
        E2.1 & 0.5057$\uparrow$ & \textbf{0.5654}$\uparrow$ & \textbf{0.9788}$\uparrow$ & 0.2574$\uparrow$ \\
        E2.2 & \textcolor{gray}{0.4895} & 0.5571 & 0.9746 & \textcolor{gray}{0.2524} \\
        E2.3 & \textbf{0.5334}$\uparrow$ & 0.5441$\downarrow$ & 0.9631$\downarrow$ & 0.3398$\uparrow$ \\
        E2.4 & 0.5125$\uparrow$ & \textcolor{gray}{0.5061}$\downarrow$ & \textcolor{gray}{0.9442}$\downarrow$ & \textbf{0.3733}$\uparrow$ \\
        E2.5 & 0.5315$\uparrow$ & 0.5308$\downarrow$ & 0.9515$\downarrow$ & 0.3709$\uparrow$ \\
        E2.6 & 0.5009$\uparrow$ & 0.5240$\downarrow$ & 0.9592$\downarrow$ & 0.2926$\uparrow$ \\
        E2.7 & 0.5319$\uparrow$ & 0.5363$\downarrow$ & 0.9585$\downarrow$ & 0.3658$\uparrow$ \\
        \bottomrule
    \end{tabular}
    \caption{Comparison of training configurations for the MF data generator. Best results are in bold, worst results are shown in gray. Arrows ($\uparrow$/$\downarrow$) indicate improvement or degradation compared to the baseline (E2.2). QWK denotes Quadratic Weighted Kappa, and Within-1 Acc denotes within-one accuracy.}
    \label{tab:mfgenerator_results}
\end{table}

\subsection{Instruction Tuning with the Value Dataset}
 We adopt a pre-trained causal language model for instruction tuning using the TULU3 supervised fine-tuning (SFT) training pipeline~\cite{lambert2024tulu3}. For the data-mixing experiments, the backbone model is Qwen3-1.7B-base~\cite{yang2025qwen3technicalreport}. For instruction tuning, we fine-tune the model on a mixture of our moral-value instruction-tuning dataset and the \texttt{allenai/tulu-3-sft-mixture} dataset~\cite{lambert2024tulu3}. 

\begin{figure*}[t]
    \centering
    \includegraphics[width=\linewidth]{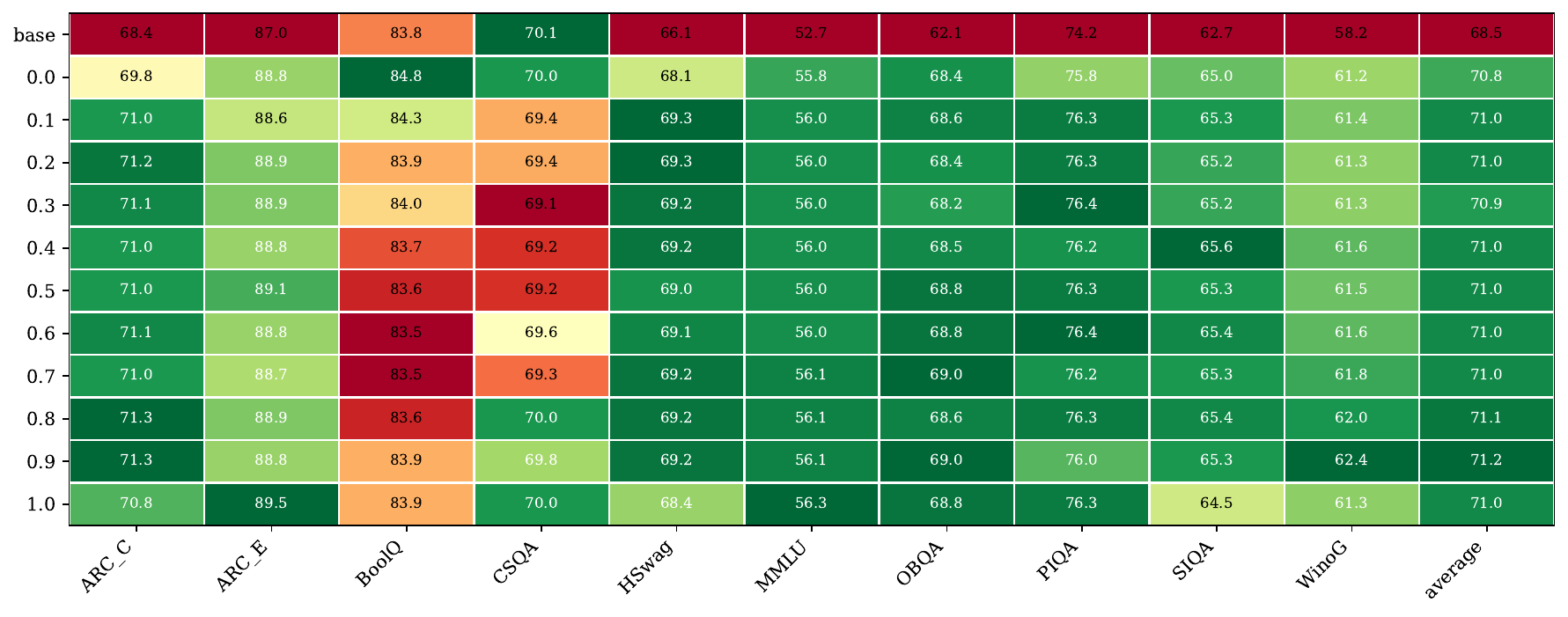}
    \caption[OLMES across mixing ratios (Qwen3-1.7B-Base)]{Performance comparison of OLMES tasks under different data mixing ratios (Qwen3-1.7B-Base).}
    \label{fig:benchmark_heatmap}
\end{figure*}

Based on the evaluation methodology introduced in Section~\ref{sec:evaluation}, we conduct experiments to investigate whether the moral-value dataset can improve value-oriented performance while maintaining general-task performance under different data mixing ratios.

Figure~\ref{fig:benchmark_heatmap} shows the models' performance on OLMES tasks. All trained models achieve significantly better performance on general tasks compared to the base model (68.5 on average). Meanwhile, varying the data mixing ratios does not lead to substantial performance differences, as all configurations yield similar average scores in the range of 70.8 to 71.2.

This result indicates that instruction tuning consistently improves the model’s ability to follow and understand natural language instructions. Notably, even when trained solely on the value dataset, the model achieves performance gains comparable to those obtained with mixtures that include general task data. This suggests that the benefits primarily stem from the instruction-tuning paradigm itself, rather than the specific composition of the training data. Our findings are aligned with those reported in FLAN~\cite{WeiBZGYLDDL22}, which demonstrate that instruction tuning enhances model performance on both few-shot and zero-shot tasks. Furthermore, the lack of significant performance differences across data mixing ratios indicates that, for a relatively small model such as Qwen3-1.7B-Base, the primary source of improvement in the first training epoch is the introduction of instruction signals, rather than the specific type or diversity of datasets. In other words, instruction tuning serves as the dominant factor driving performance gains at this scale, while dataset composition plays a comparatively minor role.

\begin{figure}[!t]
    \centering
    \includegraphics[width=0.9\linewidth]{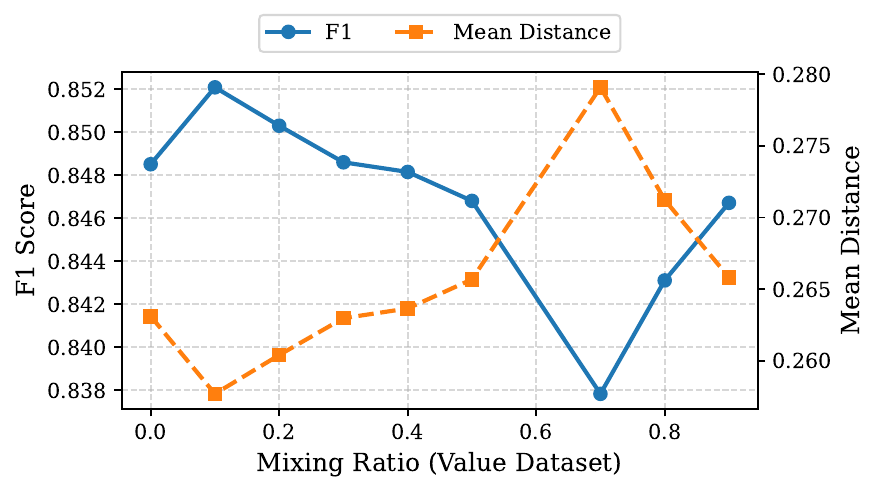}
    \caption[VIA across mixing ratios (Qwen3-1.7B-Base)]{Performance comparison of value-action gap under different data mixing ratios (Qwen3-1.7B-Base).}
    \label{fig:mixing_ratio_qwen1.7B}
\end{figure}

For value tasks, Figure~\ref{fig:mixing_ratio_qwen1.7B} presents both the F1 scores and the mean distance for models trained under different data mixing ratios. For the F1 computation, we define a binary label based on the direction of the value-action gap between Task 1 and Task 2: if both tasks exhibit the same inclination, the prediction is considered positive; otherwise, it is considered negative. The mean distance is defined as the average absolute difference between Task 1 and Task 2. Each task is binarized such that 0 denotes disagreement and 1 denotes agreement. Under this definition, the pairwise distance is either 0 (when both tasks share the same label, i.e., both agree or both disagree) or 1 (when one agrees and the other disagrees). 
We were unable to evaluate the base model and the model trained on the 100\% value dataset because neither produced outputs in the correct format for Task 1. 
For the remaining models, the F1 score increases slightly from 0.8485 with 0\% value data to a peak of 0.8521 at 10\%, after which performance gradually declines to 0.8486 at 40\%. Beyond this point, all configurations underperform the 0\% baseline, reaching the lowest F1 score of 0.8378 at 70\%. Although there is a minor recovery afterward, performance remains below that of the 0\% setting.

These results suggest two key observations. First, general-task datasets are crucial for enabling the model to understand task structure and output format, as evidenced by the failure of both the base model and the 100\% value-only model. Second, while incorporating value-oriented data improves performance on value alignment, excessive emphasis on such data appears to impair the model’s ability to follow task instructions effectively. 
Overall, a balanced mixture, specifically, a small proportion of value data (e.g., 10\%) combined with a larger amount of general task data, yields the best performance for a small model such as Qwen3-1.7B-Base on value alignment tasks, though the absolute differences across non-degenerate mixing ratios are modest.

\section{Conclusion}

We construct a value-oriented instruction-tuning dataset by first collecting and integrating three moral-value datasets, then designing prompting strategies to generate instruction–response templates and combine them with the unified dataset. The resulting dataset can be integrated into instruction-tuning pipelines and serves as a useful resource for future research on value alignment. 
For example, it can be divided into subsets based on distinct value frameworks or value types to examine how these variations affect model behavior.

There are several limitations of our current dataset. First, although we collected 421,662 samples from three datasets, our coverage remains limited to normative ethics and Moral Foundations Theory. Our pipeline can be extended to incorporate additional ethical frameworks and datasets such as valueML~\cite{scharfbillig} that based on Schwarz Value Theory~\cite{schwartz1992universals} in future work. Moreover, since our unified dataset is constructed from existing value-oriented datasets, it inevitably inherits biases present in the original sources. For example, the source datasets are primarily written in English, resulting in limited coverage of other languages and cultural perspectives. The labels generated by the data generators may also introduce noise into the unified dataset due to prediction errors. Finally, while we believe our dataset to be a useful resource, our initial experimental results should not be viewed as conclusive. Due to resource constraints, the downstream training results reported here are based on a single model family and a single run per configuration, and the influence of the dataset on larger models and across seeds remains open.

\bibliographystyle{named}
\bibliography{ijcai26_3}

\end{document}